\documentclass{IEEEtran}
\IEEEoverridecommandlockouts

\usepackage{tabularx}

\usepackage{cite}
\usepackage{amsmath,amssymb,amsfonts}
\usepackage{graphicx}
\usepackage{float}
\usepackage{algorithmic}
\usepackage[linesnumbered,ruled]{algorithm2e}
\usepackage[table]{xcolor}
\usepackage{cleveref}
\usepackage[framemethod=tikz]{mdframed}
\usepackage{multirow}
\usepackage{rotating}
\usepackage[font=footnotesize]{caption}
\usepackage{subcaption}
\usepackage{pifont} 
\usepackage{textcomp}
\usepackage{wrapfig}
\usepackage[flushmargin]{footmisc}

\newcommand{\xmark}{\ding{55}}

\SetKwRepeat{Do}{do}{while}
\let\oldnl\nl
\newcommand{\nonl}{\renewcommand{\nl}{\let\nl\oldnl}} 

\begin{document}

\title{FedSiKD: Clients \textbf{Si}milarity and \textbf{K}nowledge\textbf{D}istillation: Addressing Non-i.i.d. and Constraints in \textbf{Fed}erated Learning}

\author{
    \IEEEauthorblockN{
        Yousef Alsenani, 
        Rahul Mishra, IEEE Member, 
        Khaled R. Ahmed, 
        Atta Ur Rahman
    }
    \thanks{This work is supported by Lun Startup Studio, Riyadh 11543, Saudi Arabia. Yousef Alsenani is with the Faculty of Computing and Information Technology, King Abdulaziz University, Jeddah 21589, Saudi Arabia and R\&D Lun Startup Studio, Riyadh 11543, Saudi Arabia (e-mail: yalsenani@kau.edu.sa).}
    \thanks{Rahul Mishra is with the Dhirubhai Ambani Institute of Information and Communication Technology, Gandhinagar, Gujarat, India (e-mail: rahulmishra.rs.cse17@itbhu.ac.in).}
    \thanks{Khaled R. Ahmed is with the School of Computing, Southern Illinois University at Carbondale, IL, 62901 USA (e-mail: khaled.ahmed@siu.edu).}
    \thanks{Atta Ur Rahman is with R\&D Lun Startup Studio, Riyadh 11543, Saudi Arabia, and Riphah Institute of System Engineering, Riphah International University Islamabad, Islamabad, 46000, Pakistan (e-mail: dr.atta@lun.sa).}
    \thanks{This work has been submitted to the IEEE for possible publication. Copyright may be transferred without notice, after which this version may no longer be accessible.}
    
}

\maketitle

\begin{abstract}

In recent years, federated learning (FL) has emerged as a promising technique for training machine learning models in a decentralized manner while also preserving data privacy. 
The non-independent and identically distributed (non-i.i.d.) nature of client data, coupled with constraints on client or edge devices, presents significant challenges in FL. Furthermore, learning across a high number of communication rounds can be risky and potentially unsafe for model exploitation. Traditional FL approaches may suffer from these challenges. Therefore, we introduce FedSiKD, which incorporates knowledge distillation (KD) within a similarity-based federated learning framework. As clients join the system, they securely share relevant statistics about their data distribution, promoting intra-cluster homogeneity.
This enhances optimization efficiency and accelerates the learning process, effectively transferring knowledge between teacher and student models and addressing device constraints.
FedSiKD outperforms state-of-the-art algorithms by achieving higher accuracy, exceeding by 25\% and 18\% for highly skewed data at $\alpha = {0.1,0.5}$ on the HAR and MNIST datasets, respectively. Its faster convergence is illustrated by a 17\% and 20\% increase in accuracy within the first five rounds on the HAR and MNIST datasets, respectively, highlighting its early-stage learning proficiency. Code is publicly available and hosted on GitHub \footnote{https://github.com/SimuEnv/FedSiKD}

\end{abstract}


\section{Introduction}


Federated learning is a new and promising paradigm that has attracted a lot of interest recently in a number of fields and applications, such as smart transportation and healthcare. Federated learning enables different parties to collaborate  without exchanging private data while training their models locally. Because federated learning protects privacy, it has attracted a lot of interest. These days, a lot of commercial projects are aiming to develop and use this technology because of its bright future. Federated learning infrastructure projects are beginning to emerge, including FedML\footnote{https://www.fedml.ai/} and EleeN\footnote{https://eleenai.com/}. Other approaches are being developed by others, such as Flower\footnote{https://flower.dev/}, to train large-scale language models (LLMs).


Traditional federated learning, however, faces convergence performance issues due to the Non-i.i.d. nature of client data \cite{comm}. Several research works have demonstrated that when there is significant client variety, the FedAvg algorithm loses accuracy \cite{noniid2}. The system can experience decreases of up to 51.31\% in test accuracy and performance in cases with highly skewed data distributions \cite{noniid}. In particular, non-i.i.d. data can significantly increase this divergence because of the increased variation in the data distribution, whereas IID data across clients maintains low divergence in weights for each client \cite{noniid}. Furthermore, the resources of client devices—such as sensors and mobile phones—are usually limited, which reduces their ability to do local training \cite{imteaj}. These training procedures use millions of deep learning algorithms.


Addressing Non-i.i.d issues can be categorized into three approaches: data-based, system-based, and algorithm-based \cite{srviid}. The data-based approach involves sharing portions of datasets among clients to overcome convergence issues caused by data diversity \cite{share1, share2}. Though these initial steps significantly enhance accuracy (for example, by 30\% on the CIFAR-10 dataset), they breach the privacy preservation principle that motivates federated learning. System-based approaches, like client clustering, are other methods to reduce the diversity of client data. These approaches involve clustering clients based on similarities in loss values or model weights. The similarity-based approach, where the global server creates multiple cluster models instead of a single model, sends these cluster models to clients for training to achieve the smallest loss value. Then, these updated models are sent back to the global server \cite{sim0,sim1,sim2}. However, incorrect clustering can lead to negative knowledge transfer and degrade the system's performance

Well-known federated learning algorithms like Fedprox\cite{Fedprox}, Scaffold\cite{Scaffold}, and FedIR\cite{FedIR} are examples of algorithm-based solutions. Local model updates are guided by these methods to better match local and global optimization goals. But before they reach meaningfully high levels of accuracy, a lot more communication rounds are needed. Because of network issues or bandwidth restrictions, this may not be feasible in real-world applications. Furthermore, more communication between clients and the server raises the possibility that possible adversaries will take advantage of it. An method called knowledge distillation makes it possible to transfer information from a larger model, called a teacher, to a smaller model, called the student, in an efficient manner. Originally developed for privacy preservation\cite{KD0}, transfer learning in knowledge distillation has been used to several federated learning problems.  A primary motivation for using KD in FL was to mitigate local drift on the client side\cite{KD1, KD2}. It also tackles current resource-constrained challenges in federated learning, like communication overhead and convergence issues. However, effectively applying KD remains a complex area within machine learning\cite{kd3}. Participants in FL require a comprehensive understanding of dataset distributions and models, especially before training.


In this paper, we address the non-independent and identically distributed (non-i.i.d) and resource constraints in federated learning. We introduce FedSiKD, a client similarity and knowledge distribution in federated learning. First, clients securely share the data distribution statistics to the global server before starting the FL optimization. Second, the global server forms the clients into a number of clusters based on this received information. Finally, a federated learning technique that incorporates knowledge distillation is introduced to enable effective information exchange between instructor and student models on limited hardware.

The following is a summary of this article's contributions:
\begin{enumerate}

   \item Clients securely share dataset distribution informaiton, such as mean, standard deviation, and skewness, before joining the system

\item These shared data distribution statistics are used to create intra-cluster homogeneity, to address the clients local drift.

\item Our method integrates knowledge distillation, enabling efficient knowledge transfer between teacher and student models in each cluster. This approach mitigates resource constraints on client devices, reducing computational load by avoiding the need to run full layers of deep learning models.

\item Federated learning rounds start after clustering stage and the establishment of knowledge distillation within each cluster.

\item We demonstrate the effectiveness of our proposed system through extensive experimental results, comparing it with state-of-the-art algorithms.

\end{enumerate}

\section{Related Work}

Several methodologies have been investigated to address non-i.i.d. challenges in federated learning. These methods include data-based, system-based, and algorithm-based approaches\cite{srviid}. In the algorithm-based direction, we focus on research that incorporates knowledge distillation into the solution. Given that our methodology aligns with these three categories, we provide an in-depth discussion comparing related works to FedSiKD, justifying its significance in addressing the stated challenges.

\subsection{Data-based approach}

The data-based approach allows participants to share information about their dataset or hardware application. Rahul et al. \cite{share3} propose a resource-aware cluster in FL. Clients share their hardware information, such as CPU and memory specifications. Then, the global server clusters them into different clusters and performs the slave-master technique. Other studies allow clients to share a portion of the dataset to mitigate convergence. Zhao et al.\cite{share1} proposed a data-sharing strategy that creates a small subset of data that can be shared globally to improve training on Non-i.i.d among clients. Tuor et al.\cite{share4} proposed a federated learning protocol to overcome noisy data by allowing participants to exchange a small amount of meta-information with the global server. These protocols help clients drop out any irrelevant data before training. Our work shares a similar philosophy, where clients can initially share information with global servers. However, instead of sharing clients' data, which violates data privacy, our work allows participants to share only clients' data distribution statistics. In this direction, we still preserve the privacy that FL motivates.

\subsection{Similarity-Based Client Clustering Approach}

In this approach, studies cluster clients based on loss values or model weights. The global server and clients exchange updated models, and then the server clusters these clients based on their similarity scores derived from these values. This helps local models among clients to overcome local drift \cite{pflcluster1, pflcluster0}. Mahdi et al. \cite{sim3} proposed a method to cluster clients based on inference results. In this method, the global server receives the updated model from each client and captures the inference results. These results are used to form the adjacency matrix, which then facilitates the identification of clusters. Li et al.\cite{sim4} employed two novel measurements to cluster clients, utilizing the Monte Carlo method and Expectation Maximization under the assumption of a Gaussian Mixture Model. These cluster protocols target a Peer-to-Peer framework. Although clustering can expedite convergence, it might mislead the optimization if the clusters are poorly constructed. In contrast to current approaches that cluster based on loss values and model parameters, in this paper, we cluster based on client dataset statistics, where clients initially share these statistics before optimization.

\subsection{Knowledge Distillation and Transfer in Federated Learning}

In this section, we discuss research that employs Knowledge Distillation (KD) to address local client drift, resource constraints, heterogeneity, and to achieve personalized FL.

\subsubsection{Clients resource limitation}

A set of techniques\cite{6,7,10} are presented that allow user training of much smaller models than the edge server, exploiting KD as an interchange protocol among model representations, thus decreasing on-device computing costs in federated edge learning (FEL). 
In particular, works \cite{6,10} develop FEL approaches that utilize alternating minimization methods. These works used knowledge transfer from compact on-device models to a larger edge model using knowledge distillation (KD).
Subsequently, the on-device models are optimized based on the knowledge transferred back from the edge.
Moreover, the authors in\cite{7} introduced an ensemble knowledge transfer model that enables collaborative training among devices. In this work, the large edge model learns a weighted consensus of the uploaded small on-device models through KD.
The author in\cite{FD} introduced a novel training approach that enables data-independent knowledge transfer by utilizing a distributed generative adversarial network (GAN). This approach generates shared space representations that can be effectively utilized in federated distillation.

\subsubsection{Heterogeneous Resources}

Federated edge learning (FEL) encompasses significant resource heterogeneity, including hardware and operating system configurations as well as network connectivity. These challenges require KD-based FEL approaches to adapt the training and learning processes to accommodate power and hardware heterogeneity and to tolerate dynamic environments. The heterogeneity can cause up to 4.6x and 2.2x degradation in the quality and performance\cite{HereImpct}. 
Instead of aggregating the weights of each local model from each client, the approach in\cite{KDExch} is to aggregate the local knowledge extracted from these clients to construct global knowledge through the process of knowledge distillation.
To mitigate the heterogeneity of clients' devices, an approach\cite{Fedbkd} can be adopted where an autoencoder is used to generate a synthetic dataset on a cloud server. This synthetic dataset is made available for clients to utilize, eliminating the need for them to upload their private datasets. The synthetic dataset is designed to have similar classes as the clients' private datasets. In this approach, a bidirectional knowledge distillation technique is employed to facilitate learning between the clients and the cloud.

\subsubsection{Personalization}

Knowledge distillation and personalized federated learning functioned effectively to overcome a number of federated learning issues, such as data heterogeneity and communications overhead\cite{pFL0,pFL1}.
The authors in\cite{FedICT} propose a method to adopt the local and global knowledge exchange between clients and servers in a multiaccess edge environment. The objective of this work is to address the Non-i.i.d. nature and data drift among clients through personalized optimization. The authors of \cite{pFL2} suggested a meta-federated learning strategy in the healthcare sector that utilized knowledge distillation. The author conducted the experiment using different data collection degrees and achieved around 10\% improvement.   . The objective of this work is to address the Non-i.i.d. nature and data drift among clients through personalized optimization.

In our study, we aim to address challenges related to the non-i.i.d. nature of clients' datasets and resource constraints challenges in federated learning. The majority of studies customize and personalize the local model during optimization and run standard deep learning models at the clients' local end. The direct application of these state-of-the-art methods might not yield satisfactory results in addressing these challenges, as shown in Table,attached to this paper. Our novel framework allows clients to share dataset statistics instead of actual raw data to preserve privacy. This proposed step enhances performance and can result in safer and faster optimization, avoiding a high number of communications. Finally, we tackle resource limitations with teacher and student models, thereby avoiding the need to run full layers of deep learning at the clients' local end.

\begin{table}[h]
\centering
\caption{Comparisons With State-of-Art}
\label{table:comparison}
\begin{tabular}{p{2cm} p{1cm} p{2cm} p{2cm}}
\hline
Method & Privacy & Knowledge-Distillation  & Fast Convergence \\
\hline
\cite{Fedprox},\cite{Scaffold},\cite{FedIR}& \checkmark & \xmark & \xmark  \\
\cite{share1, share2} & \xmark & \xmark & \checkmark \\
\cite{fast1, fast2} & \checkmark & \xmark & \checkmark \\
\cite{6,7,FD} & \checkmark & \checkmark & \xmark  \\
FedSiKD & \checkmark & \checkmark & \checkmark  \\
\hline
\end{tabular}
\end{table}

\begin{table}[ht]
\centering
\caption{Summary of Notations}
\label{tab:notations}
\begin{tabular}{c l}
\hline
\textbf{Notation} & \textbf{Description} \\
\hline
\( C \) & Set of all clients in the federated learning system \\
\( N \) & Total number of clients \\
\( c_i \) & The \( i \)-th client \\
\( Q \) & Set containing all classes in the classification problem \\
\( q \) & Total number of classes \\
\( D_i \) & Local dataset of client \( c_i \) \\
\( d_i \) & Number of instances in dataset \( D_i \) \\
\( x_{ij} \) & The \( j \)-th instance in dataset \( D_i \) \\
\( y_{ij} \) & Class label corresponding to instance \( x_{ij} \) \\
\( \Pi_i \) & Classifier for client \( c_i \) \\
\( B_i \) & Batch size used for training on client \( c_i \) \\
\( \tau_i \) & Number of SGD operations in one training round on client \( c_i \) \\
\( E \) & Number of local epochs needed for training on client \( c_i \) \\
\( \theta_1, \theta_2 \) & Local parameters for two distinct clients \\
\( wc \) & Parameters of an ideal centralized model \\
\( wf \) & Parameters obtained from FedAvg \\
\( h_1, h_2 \) & Local drift for the clients \\
\( K \) & Number of clusters \\
\( C_k \) & Set of clients in cluster \( k \) \\
\( T_{t-1}^{c(k)} \) & Teacher model for cluster \( c(k) \) at time \( t-1 \) \\
\( S_{t-1}^{c(k),i} \) & Student models within cluster \( c(k) \) at time \( t-1 \) \\
\( w_{t-1}^g \) & Global model weights from previous round \\
\( w_{t-1}^{c(k)} \) & Local model weights for cluster \( c(k) \) \\
\( \bar{w}_t^{c(k)} \) & Average model weights for cluster \( c(k) \) after training \\
\( w_t^g \) & Updated global model weights \\
\( \theta_i^t \) & Local model parameters for client \( i \) at time \( t \) \\
\( \theta_i^* \) & Optimal model parameters for client \( i \) \\
\( \bar{\theta}_t \) & Average model parameters across all clients at time \( t \) \\
\( Var_{intra} \) & Variance within a cluster \\
\( Var_{total} \) & Total variance across all clusters \\
\hline
\end{tabular}
\end{table}

\section{Preliminaries}

This study discuses a federated learning system involving a collection \(C\) of \(N\) clients and a global federated server, where \(C = \{c_1, \ldots, c_N\}\). The focus is on a multi-class classification problem within a set \(Q\) containing \(q\) classes, denoted as \(Q = \{1, \ldots, q\}\). Each client \(c_i\) possesses a local dataset \(D_i\) with \(d_i\) instances and a subset of classes from \(Q\), where \(1 \leq i \leq N\). Instances in \(D_i\) are represented as pairs \((x_{ij}, y_{ij})\), for \(1 \leq j \leq d_i\). The objective when training the model on client \(c_i\) is to learn the association between \(x_{ij}\) and \(y_{ij}\) for all instances, \(\forall j \in \{1 \leq j \leq d_i\}\), to construct a classifier \(\Pi_i\) that predicts class labels for unknown instances during testing. The batch size used for training on client \(c_i\) is \(B_i\), and the number of Stochastic Gradient Descent (SGD) operations in one training round on \(c_i\) is denoted as \(\tau_i\). The estimation of \(\tau_i\) is given by \(\tau_i = \left\lfloor \frac{E \cdot d_i}{B_i} \right\rfloor\), where \(E\) represents the number of local epochs needed for training on \(c_i\).

Our federated learning system comprises two primary components: clients and a central server. The clients are devices with local and private data and constrained computational capabilities. They are responsible for training local student models using teacher models obtained from a leader clients. We assume that these leader clients, functioning as intermediary nodes, possess greater computational power and assist the learning process by aggregating client models, conducting knowledge distillation, and compressing teacher models before distributing them to the clients. The central server manage the entire process and may maintain the global model.

\begin{figure}
		\centering
		\includegraphics[width=1\columnwidth,keepaspectratio]{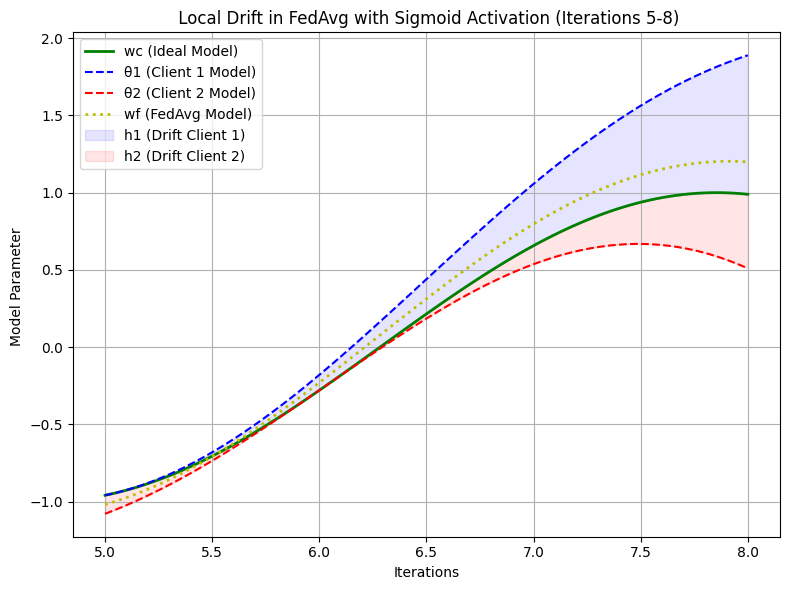}
		\caption{Local drift}
		\label{fig:dreft}
	\end{figure}
 
\subsection{Non-i.i.d. in Federated Learning}

Federated Learning (FL) fundamentally involves training decentralized models on local datasets. When these datasets exhibit Non-i.i.d. (Non-Independent and Identically Distributed) characteristics, a phenomenon known as 'local drift' can occur. Such discrepancies in data distributions can skew the global model. A significant challenge arises with the Federated Averaging (FedAvg) algorithm, which may become less efficient in the presence of Non-i.i.d. data, leading to deviations from the ideal global model.

To illustrate, consider Figure \ref{fig:dreft}, which demonstrates the application of a non-linear transformation function \( f \), such as a Sigmoid activation. Let \(\theta_1\) and \(\theta_2\) be the local parameters of two distinct clients, with \( wc \) representing the parameters of an ideal centralized model, and \( wf \) representing the parameters obtained from FedAvg \cite{noniid}.

The local drift for each client is denoted as:
\[ h_1 = wc - \theta_1 \]
\[ h_2 = wc - \theta_2 \]
For a given data point \( x \), the predictions made by the clients' models are:
\[ y_1 = f(\theta_1, x) \]
\[ y_2 = f(\theta_2, x) \]
FedAvg computes the averaged parameter as:
\[ wf = \frac{\theta_1 + \theta_2}{2} \]
Ideally, if the function \( f \) were linear, one would expect that applying \( f \) to \( wf \) would yield the average of the clients' predictions:
\[ f(wf, x) \approx \frac{y_1 + y_2}{2} \]
However, due to the non-linearity of \( f \), this is not the case, and we observe:
\[ f(wf, x) \neq \frac{y_1 + y_2}{2} \]
This discrepancy underscores the impact of non-linear transformations on the aggregation process in FedAvg, leading to a global model (\( wf \)) that may not accurately represent the combined knowledge of the local models, especially in the presence of Non-i.i.d. data.

\begin{figure}
		\centering
		\includegraphics[width=1.11\columnwidth,keepaspectratio]{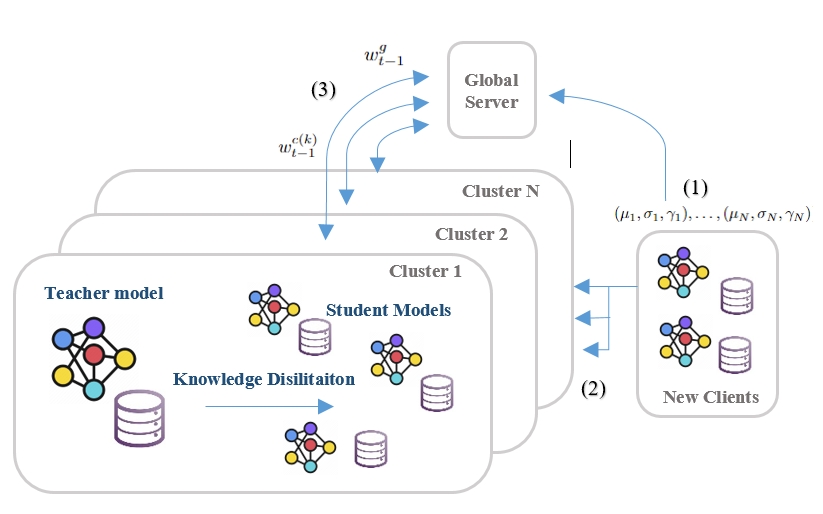}
		\caption{(1) Clients share their statistics with the global server to identify an appropriate cluster.
(2) The global server assigns clients to clusters.
(3) Knowledge distillation and federated learning training proceed within each cluster.}
		\label{fig:KD}
	\end{figure}
\section{FedSiKD} 
In this section, we introduce FedSiKD Clients Similarity and Knowledge Distillation: Addressing Non-i.i.d. and Constraints in Federated Learning. FedSiKD consists of several distinct phases. First, we discuss the initial information that clients share about their dataset statistics, including mean, standard deviation, and skewness. Then, we illustrate how we form clusters based on this information shared by FL clients. The Knowledge distillation is introduced in detail, including aspects of teacher and student models. Finally, we present the federated learning optimization process.

\subsection{Similarity-Based Client Clustering}

Before the optimization phase in federated learning, clients are asked to share key statistical measures of their local data with the central server. This study assumes differential privacy is applied to this shared information, although developing an exact differential privacy model is beyond the scope of this paper. This step is instrumental in tackling the challenges posed by Non-i.i.d. data distributions. The shared statistics typically include the mean \( \mu_i \), standard deviation \( \sigma_i \), and skewness \( \gamma_i \) of each client's dataset, represented as:

\begin{equation}
\text{client\_stats} = \{ (\mu_1, \sigma_1, \gamma_1), \ldots, (\mu_N, \sigma_N, \gamma_N) \}
\end{equation}

Now, after receiving these statistical information, global server can cluster the clients. Now, after receiving these statistical information, global server can cluster the clients. To find the optimal number of clusters \( K \), the server employs metrics such as the Silhouette coefficient\cite{SC}, the Calinski-Harabasz Index\cite{CH}, and the Davies-Bouldin Index\cite{DC}. Global server utilize these metrics to assess the quality of clustering and determining the the appropriate \( K \).

The  global sever apply k-means clustering algorithm, to minimize the variance within cluster:

\begin{equation}
J = \sum_{k=1}^{K} \sum_{x_i \in C_k} \| x_i - \mu_k \|^2
\end{equation}

The number of clusters, and the set of clients in cluster are denoted by \( K \), and \( C_k \), respectively. \( x_i \),  \( \mu_k \) are  the statistical measures for client \( i \)  and the centroid of cluster \( k \) respectively.
The algorithm is continuing resigning clients to cluster and updates centroids towards the optimal configuration. Each client \( i \) is thereby assigned to a unique cluster \( c(i) \), where \( 1 \leq c(i) \leq K \).

\subsection{Convergence and Complexity Analysis}

\textbf{Theorem 1:} For a given client's distribution $p_i$, it holds that: Given the client clustering procedure based on data distribution characteristics $\mu_i, \sigma_i,$ and $\gamma_i$ (mean, standard deviation, and skewness), under the assumptions that the loss function is convex and the clusters are well-separated

\begin{equation}
E\left[\lVert \theta_i^t - \theta_i^* \rVert^2\right] \leq E\left[\lVert \bar{\theta}_t - \theta_i^* \rVert^2\right]
\end{equation}

\textbf{Proof:}

The proof relies on the following assumptions:

\begin{itemize}
    \item The loss function used to train the models is convex with respect to the model parameters.
    \item The data within each cluster follows a distribution that is more similar among the clients within the cluster than across clusters, i.e., the clusters are well-separated in terms of the data distribution characteristics.
\end{itemize}

\begin{enumerate}
    \item \textit{Cluster Homogeneity:} \\
    By the given clustering procedure, clients within the same cluster have data distributions characterized by closer mean $\mu$, standard deviation $\sigma$, and skewness $\gamma$. Hence, the gradient updates from each client in the same cluster are more similar, reducing the variance of the parameter updates within the cluster.

    \item \textit{Optimal Cluster Parameter:} \\
    For each cluster, we define an optimal parameter $\theta^*_i$ that minimizes the expected loss over the data distributions within that cluster. Due to the homogeneity within the cluster, the sequence of model parameters $\{\theta_i^t\}$ generated by each client's local updates is more likely to converge to $\theta^*_i$ than to $\bar{\theta}_t$, which is the average of parameters across potentially heterogenous clusters.

    \item \textit{Bounding the Error:} \\
    The error between the local parameters $\theta^t_i$ and the optimal $\theta^*_i$ can be bounded by the variance within the cluster, denoted as $Var_{intra}$. Conversely, the error for $\bar{\theta}_t$ is affected by the overall variance across all clusters, denoted as $Var_{total}$.

    We define $Var_{intra}$ and $Var_{total}$ as follows:
    \begin{align*}
    Var_{intra} &= \frac{1}{|C_k|} \sum_{\theta_i \in C_k} \lVert \theta_i - \mu_k \rVert^2, \\
    Var_{total} &= \frac{1}{N} \sum_{i=1}^{N} \lVert \theta_i - \bar{\theta} \rVert^2,
    \end{align*}
    where $C_k$ is the set of client indices in cluster $k$, $\mu_k$ is the centroid of cluster $k$, and $\bar{\theta}$ is the global average of the parameters.

    Given the assumption of well-separated clusters, we have:
    \begin{equation}
    Var_{intra} \leq Var_{total}
    \end{equation}

    Consequently, it follows that the expected squared distance between the local parameters and the optimal parameters is less than or equal to that of the average parameters across all clients:
    \begin{equation}
    E\left[ \lVert \theta^t_i - \theta^*_i \rVert^2 \right] \leq E\left[ \lVert \bar{\theta}_t - \theta^*_i \rVert^2 \right]
    \end{equation}
\end{enumerate}

\textbf{Theorem 2:} Under the assumptions of Theorem 1,  FedSiKD demonstrates fast convergences.

The initial parameters $\theta_0^i$ for each client $i$ can be within
$\delta$-neighborhood of the optimal parameters $\theta^*$. For all
clients, the difference within the model can be less than $\delta$-neighborhood as $\lVert \theta_0^i - \theta^* \rVert \leq \delta$
which can be close to the optimal goal. For the gradients $\nabla F_c(\theta)$ for all clients $k$ in a cluster are similar.
We can measure the convergence rate by

\begin{equation}
\rho_t = \min_{c \in [C]} \Big\{ \eta_t \lambda_{\min} (\nabla^2 F_c(\theta_t)) \Big\}
\end{equation}
where $\rho_t$ is convergence rate at each round $t$ as:

where $\lambda_{\min}(\cdot)$ is the lowest value, showing the curve of the loss function, and $\eta_t$ represents the learning rate at round $t$. The proper step size toward the minimum can be measured by the convergence factor $\rho_t$.

Considering the cluster-based initialization and the bounded gradient dissimilarity within clusters, we establish that:

\begin{equation}
E\left[\lVert \theta_t - \theta^* \rVert^2\right] \leq (1-\rho_t)^t \lVert \theta_0 - \theta^* \rVert^2 + \frac{\sigma^2}{2\mu} \sum_{\tau=1}^{t} (1-\rho_{\tau})^{t-\tau} \eta_{\tau}^2
\end{equation}

where $\sigma^2$ denotes the variance of stochastic gradients, and $\mu$ is the strong convexity constant. The term $\frac{\sigma^2}{2\mu} \sum_{\tau=1}^{t} (1-\rho_{\tau})^{t-\tau} \eta_{\tau}^2$ quantifies the cumulative effect of this gradient variability over multiple training rounds in the federated learning process.

\subsection{Knowledge Distillation}

Building upon the client clustering framework and its associated convergence properties, we now explore an advanced strategy to leverage the homogeneity within clusters for enhanced model training. This involves applying knowledge distillation techniques within the confines of our established clusters.

\subsubsection{Clustered Knowledge Distillation}

Given the clusters \( C_k \) for \( k = 1, 2, \ldots, K \) derived from the data distributions and client statistics we discuss earlier, there is an inherent advantage in utilizing the collective knowledge within these clusters. Each cluster can be thought of as a localized knowledge domain, where the data distributions \( D(x_i, x_j) \) are relatively homogeneous.

\subsubsection{The Teacher-Student Paradigm}

For every cluster \( C_k \), we assume that a leader client model can act as the 'teacher', denoted by \( T_{t-1}^{c(k)} \). The teacher client within the cluster is selected based on various criteria, such as the amount of available computing resources for training. The optimal selection is beyond the scope of this paper.
Given the minimized intra-cluster distances and the findings from our convergence and complexity analysis, particularly the properties related to local model convergence \( h_{ij} \), this teacher model embodies the key characteristics of its cluster's data distribution. Subsequent client models within \( C_k \) act as'students', each denoted \( S_{t-1}^{c(k),i} \) for \( i = 1, 2, \ldots, |c(k)| - 1 \).

Knowledge Distillation facilitates the transfer of insights from the teacher model \( T_{t-1}^{c(k)} \) to the student model \( S_{t-1}^{c(k),i} \). By doing so, each student model not only refines its parameters based on its own client data but also integrates knowledge representative of the entire cluster \( C_k \). This mechanism ensures that the enhanced local models, when aggregated, make significant contributions towards an enriched global model.
For each communication round \( t \), we iterate through the clusters and create a local model for each cluster. Let \( w_{t-1}^g \) denote the global model weights obtained from the previous round. We initialize the local models with the global weights:

\[ w_{t-1}^{c(k)} = w_{t-1}^g \quad \text{for } k = 1, 2, \ldots, K \]


The training process involves the following steps:

\subsubsection{Teacher Training}

The teacher model is trained as follows:

\[
T_t^{c(k)} = \text{{argmin}} \sum_{x_i \in c(k)} \text{{Loss}}(T_{t-1}^{c(k)}(x_i), y_i)
\]

\subsubsection{Student Training}

We train the student models using the distillation loss. The student models are trained as follows:

\begin{align*}
S_t^{c(k),i} = & \text{{argmin}} \sum_{x_i \in c(k) \setminus \{c(k,i)\}} \text{{Loss}}(S_{t-1}^{c(k),i}(x_i), y_i) \\
& + \text{{DistillationLoss}}(T_t^{c(k)}(x_i), S_{t-1}^{c(k),i}(x_i)) \\
\end{align*}

where $c(k,i)$ represents the $i$-th student in cluster $c(k)$.
\subsubsection{Clustered Federated Learning}
After training the teacher and student models, we scale the weights according to the number of clients in each cluster and combine them to obtain the average cluster weights:

\[
\bar{w}_t^{c(k)} = \frac{1}{|c(k)|} \sum_{i=1}^{|c(k)|} w_t^{c(k),i}
\]

Finally, the average cluster weights are used to update the global model weights:

\[
w_t^g = \frac{1}{K} \sum_{k=1}^{K} \bar{w}_t^{c(k)}
\]

\subsection{FedSiKD Algorithm}
In this section, we discuss the steps of FedSiKD. (lines 1-4) FedSiKD calls $ClientStatisticsSharing()$ to receive data distribution statistics from clients. Then, the algorithm (lines 5-7) calls the $ClusterFormation()$  function to form clients into clusters. The FedSiKD algorithm runs the $KnowledgeDistillation()$ function for the knowledge distillation. At each cluster, we initialize and train teacher and student models, and then we aggregate the student model weights to update the cluster weights (lines 8-16). Finally, we update the global model weights with the averaged cluster weights (line 18).

\begin{algorithm}[h]
\DontPrintSemicolon
\caption{FedSiKD: Federated Learning with Similarity-based Client Clustering and Knowledge Distillation}
\label{alg:FedSiKD}

\SetKwInput{KwInput}{Input}
\SetKwInput{KwOutput}{Output}
\SetKwInput{KwInitialize}{Initialize}
\SetKwProg{Fn}{Function}{:}{}
\SetKwFunction{FMain}{Main}
\SetKwFunction{FClientStat}{ClientStatisticsSharing}
\SetKwFunction{FClusterForm}{ClusterFormation}
\SetKwFunction{FKnowledgeDistill}{KnowledgeDistillation}

\KwInput{Client datasets $\{D_i\}_{i=1}^N$, number of clusters $K$, number of communication rounds $T$}
\KwOutput{Global model weights $w_t^g$}
\KwInitialize{Global model weights $w_0^g$}

\Fn{\FClientStat{}}{
    \For{each client $i \in \{1, \ldots, N\}$}{
        Client $i$ computes and sends $\mu_i, \sigma_i, \gamma_i$ to the server\; \nllabel{line:client_stats}
    }
}

\Fn{\FClusterForm{}}{
    The server computes the optimal number of clusters $K$ using the Silhouette coefficient, Calinski-Harabasz Index, and Davies-Bouldin Index\;
    Server applies K-means clustering to form clusters $\{C_k\}_{k=1}^K$\; \nllabel{line:cluster_formation}
}

\Fn{\FKnowledgeDistill{}}{
    \For{$t = 1$ to $T$}{ \nllabel{line:for_rounds}
        \For{each cluster $k \in \{1, \ldots, K\}$}{ \nllabel{line:for_clusters}
            Initialize a teacher model $T_{t}^{c(k)}$ and student models $\{S_{t}^{c(k),i}\}$\; \nllabel{line:init_models}
            Train teacher model $T_t^{c(k)}$ on cluster data\; \nllabel{line:train_teacher}
            \For{each student $i$ in cluster $C_k$}{ \nllabel{line:for_students}
                Train student model $S_t^{c(k),i}$ using distillation loss with $T_t^{c(k)}$\; \nllabel{line:train_student}
            }
            \tcp{Aggregate student model weights to update cluster weights}
            $\bar{w}_t^{c(k)} \gets \frac{1}{|C_k|} \sum_{i=1}^{|C_k|} w_t^{c(k),i}$\; \nllabel{line:aggregate_weights}
        }
        \tcp{Update the global model weights with the averaged cluster weights}
        $w_t^g \gets \frac{1}{K} \sum_{k=1}^{K} \bar{w}_t^{c(k)}$\; \nllabel{line:update_global}
    }
}

\Fn{\FMain{}}{
    \FClientStat{}\;
    \FClusterForm{}\;
    \FKnowledgeDistill{}\;
}

\end{algorithm}



(Anguita et al. 2013): 6-class human activity recognition dataset collected via sensors embedded in 30
users’ smartphones (21 users’ datasets used for training
and 9 for testing, and each has around 300 data samples).

\section{Experiment}

\subsection{Dataset and Models}

\subsubsection{Dataset}
This study utilizes four publicly available and well-known datasets in federated learning: MNIST\cite{minst}, HAR\cite{har}, which consist of reliable labeling. The MNIST dataset features 50,000 training images and 10,000 testing images of handwritten digits ranging from 0 to 9. The HAR dataset captures three-dimensional measurements from both an accelerometer and a gyroscope. The goal of the dataset's structure is to enable a 6-class classification issue, which uses these sensor information to anticipate user behavior with high accuracy. There are around 10,299 cases in all, and 561 characteristics characterize each one. 

To simulate different levels of Non-i.i.d. data distribution among 40 clients, we employed the Dirichlet distribution with parameters $\alpha = \{2.0, 1.0, 0.5, 0.1\}$. As $\alpha$ decreases, the label distribution becomes more heterogeneous, thus indicating a shift towards a Non-i.i.d. data distribution. We run our experiment in environment that has two 16 GB accelerator cards in addition to two dual multicore CPUs, each with a minimum of 12 cores. 

\subsubsection{Models}
In our experimental setup, we employed two distinct convolutional neural network (CNN) architectures: the Teacher Model and the Student Model. We illustrate the data structure and models for each dataset in the tables. Tables \ref{table:HAR} and \ref{table:MNIST} present the model structures used for the datasets. We ran 70 and 50 rounds on the MNIST and HAR datasets, respectively, with a batch size of 64.

\begin{table}
\centering
\caption{Model Structure For MNIST}
\label{table:MNIST}
\resizebox{\columnwidth}{!}{%
\begin{tabular}{|c|c|p{6.5cm}|}
\hline
\textbf{Layer \#} & \textbf{Model} & \textbf{Key Properties} \\
\hline
1 & \multirow{5}{*}{Teacher} & - Input: input Shape, Conv2D: 32, (3, 3), strides=(2, 2), padding="same" \\
\cline{1-1}\cline{3-3}
2 & & - Conv2D: 64, (3, 3), strides=(2, 2), padding="same" \\
\cline{1-1}\cline{3-3}
3 & & - Conv2D: 64, (3, 3), strides=(2, 2), padding="same" \\
\cline{1-1}\cline{3-3}
4 & & - Conv2D: 64, (3, 3), strides=(2, 2), padding="same", Flatten \\
\cline{1-1}\cline{3-3}
5 & & - Dense: 10, 'softmax' \\
\hline
1 & \multirow{5}{*}{Student} & - Input: inputShape, Conv2D: 32, (3, 3), strides=(2, 2), padding="same" \\
\cline{1-1}\cline{3-3}
2 & & - Conv2D: 16, (3, 3), strides=(2, 2), padding="same" \\
\cline{1-1}\cline{3-3}
3 & & - Conv2D: 16, (3, 3), strides=(2, 2), padding="same" \\
\cline{1-1}\cline{3-3}
4 & & - Conv2D: 64, (3, 3), strides=(2, 2), padding="same", Flatten \\
\cline{1-1}\cline{3-3}
5 & & - Dense: 10, 'softmax' \\
\hline
\end{tabular}%
}
\end{table}

\begin{table}
\centering
\caption{Model Structure For HAR}
\label{table:HAR}
\resizebox{\columnwidth}{!}{%
\begin{tabular}{|c|c|p{6.5cm}|}
\hline
\textbf{Layer \#} & \textbf{Model} & \textbf{Key Properties} \\
\hline
1 & \multirow{2}{*}{Teacher} &  Conv1D: 128, 3, strides=2, padding="same", LeakyReLU: 0.2, MaxPooling1D: pool\_size=2, strides=1, padding="same", Dropout: 0.25 \\
\cline{1-1}\cline{3-3}
2-5 & & Conv1D: 256, 3, strides=2, padding="same", Flatten, Dense: 128, 'relu', Dense: 6, 'softmax' \\
\hline
1 & \multirow{2}{*}{Student} &  Conv1D: 64, 3, strides=2, padding="same", LeakyReLU: 0.2, MaxPooling1D: pool\_size=2, strides=1, padding="same", Dropout: 0.25 \\
\cline{1-1}\cline{3-3}
2-5 & & Conv1D: 256, 3, strides=2, padding="same", Flatten, Dense: 128, 'relu', Dense: 6, 'softmax' \\
\hline
\end{tabular}%
}
\end{table}

\subsubsection{Baseline Algorithms}
Our analysis includes state-of-the-art algorithms, notably the \textit{FL+HC} algorithm \cite{FLHL}, which leverages hierarchical clustering based on model updates. It employs agglomerative clustering with 'average' linkage and uses Euclidean distances to group model weights. Clusters are determined by cutting the hierarchical tree at a designated distance, enabling adaptive clustering within the federated learning framework. Additionally, we compare our approach with \textit{Random Clustering}, which assigns clients to clusters randomly. Lastly, we include the well-established \textit{FedAvg} algorithm for comparison \cite{fedavg}.

\subsection{Discussion}

We conduct the experiment with different algorithms: FedSiKD, FL+HC, Random Clustering, and FedAvg, running over two datasets: MNIST, and HAR, which were developed to apply multi-class classification. We tested various levels of Non-i.i.d data distribution among 40 clients. The data and label distributions are controlled by the $\alpha$ value, where a lower value indicates a more heterogeneous distribution.

\begin{figure*}[h]
    \centering
    \begin{subfigure}[b]{0.235\textwidth}
        \includegraphics[width=\textwidth]{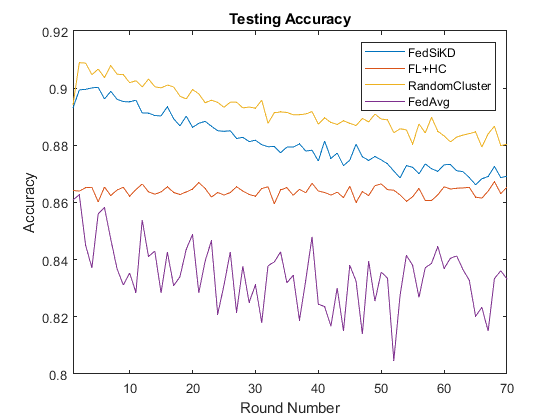}
        \caption{MNIST $\alpha$ = 2.0}
        \label{fig:mnist_alpha20_acc}
    \end{subfigure}
    ~ 
    \begin{subfigure}[b]{0.235\textwidth}
        \includegraphics[width=\textwidth]{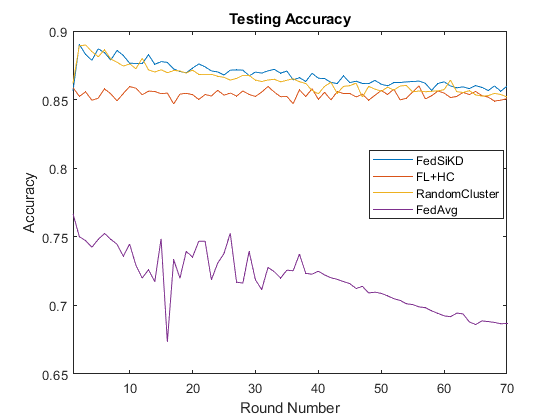}
        \caption{MNIST $\alpha$ = 1.0}
        \label{fig:mnist_alpha10_acc}
    \end{subfigure}
    ~ 
    \begin{subfigure}[b]{0.235\textwidth}
        \includegraphics[width=\textwidth]{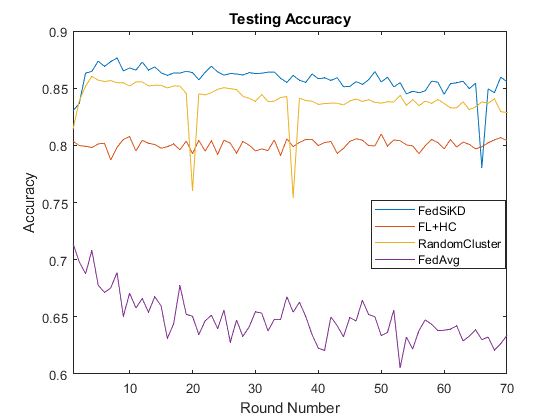}
        \caption{MNIST $\alpha$ = 0.5}
        \label{fig:mnist_alpha05_acc}
    \end{subfigure}
    ~ 
    \begin{subfigure}[b]{0.235\textwidth}
        \includegraphics[width=\textwidth]{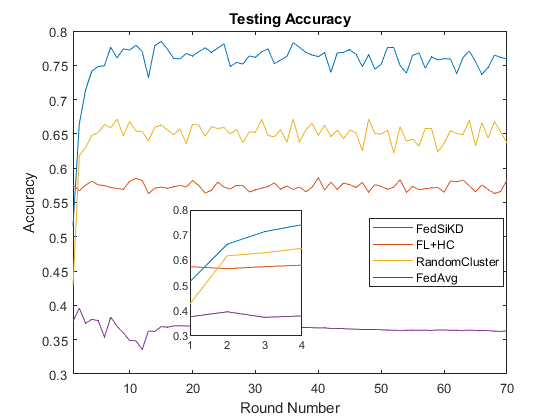}
        \caption{MNIST $\alpha$ = 0.1}
        \label{fig:mnist_alpha01_acc}
    \end{subfigure}
    
    \begin{subfigure}[b]{0.235\textwidth}
        \includegraphics[width=\textwidth]{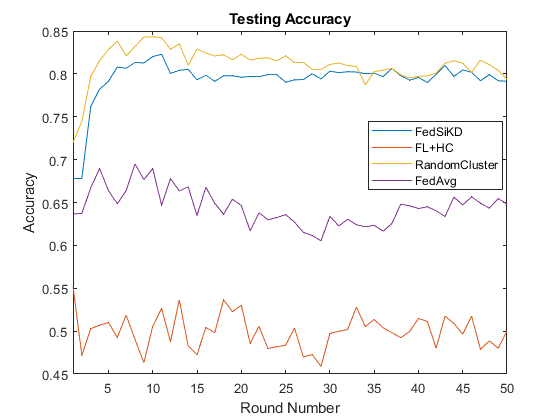}
        \caption{HAR $\alpha$ = 2.0}
        \label{fig:har_alpha20_acc}
    \end{subfigure}
    ~ 
    \begin{subfigure}[b]{0.235\textwidth}
        \includegraphics[width=\textwidth]{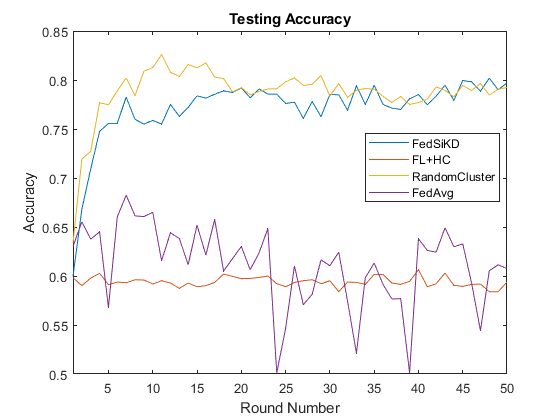}
        \caption{HAR $\alpha$ = 1.0}
        \label{fig:har_alpha10_acc}
    \end{subfigure}
    ~ 
    \begin{subfigure}[b]{0.235\textwidth}
        \includegraphics[width=\textwidth]{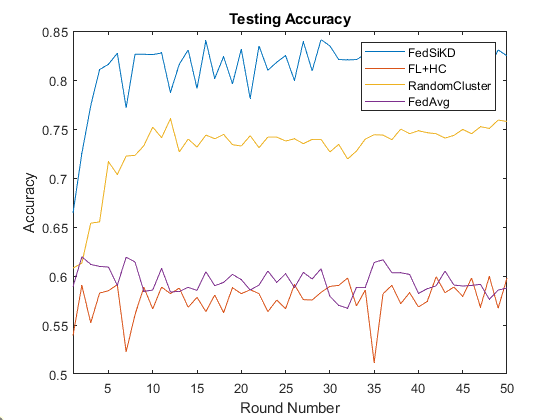}
        \caption{HAR $\alpha$ = 0.5}
        \label{fig:har_alpha05_acc}
    \end{subfigure}
    ~ 
    \begin{subfigure}[b]{0.238\textwidth}
        \includegraphics[width=\textwidth]{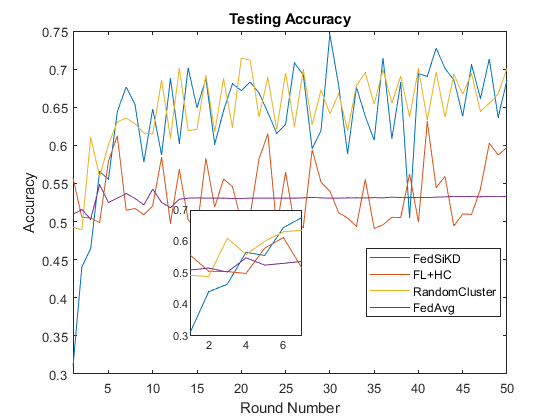}
        \caption{HAR $\alpha$ = 0.1}
        \label{fig:har_alpha01_acc}
    \end{subfigure}
    
    \caption{Test Accuracy for MNIST and HAR Datasets at Different Levels of Non-i.i.d Data Distribution}
    \label{fig:datasets_accuracy}
\end{figure*}

In Figure \ref{fig:datasets_accuracy}, the upper figures show the results for the \textbf{MNIST} dataset. At $\alpha = 2.0$, where data distribution among clients is more uniform, FedSiKD and RandomCluster exhibit similar performances, outperforming FL+HC and FedAvg. At $\alpha = 1.0$, as data becomes less uniform, FedSiKD maintains performance and outperforms all other algorithms. While RandomCluster performs slightly better with more uniform data distribution, it doesn't handle highly skewed data well, which is a real-world scenario and a main challenge in federated learning. At $\alpha = 0.5$ and $0.1$, where the data is more skewed and highly non-i.i.d., differences between algorithms are more pronounced.
Importantly, at $\alpha = 0.1$, FedSiKD shows a significant improvement in accuracy in the first four rounds, from 51\% to 74\%. This indicates FedSiKD's ability to rapidly adapt to a federated learning environment. It suggests that the initial model parameters and the learning algorithm are well-suited to quickly assimilate distributed knowledge across clients, and it could also be considered a method for quick adaptation in early rounds. Such rapid improvement indicates successful knowledge transfer from teacher to student models, especially in contexts with highly skewed data distributions. The results could be further improved by employing a more robust student model, which is beyond the scope of this research.

In Figure \ref{fig:datasets_accuracy}, the lower figures show the results for the \textbf{HAR} dataset.
The behavior of all four algorithms run on the HAR dataset is quite similar to the results obtained from the MNIST dataset. FedSiKD and RandomCluster demonstrate steady performance and higher accuracy than the others when the data becomes less heterogeneous, at \(\alpha = 2.0\). 
At $\alpha$ = 1.0, while RandomCluster initially leads among the algorithms, its accuracy decreases due to sensitivity to data distribution changes. In contrast, FedSiKD's accuracy shows a steady and consistent improvement. At \(\alpha = 0.5\), where the data becomes more skewed, FedSiKD handles the heterogeneity more adaptively compared to others. However, FedSiKD, FL+HC, and random clustering exhibit greater fluctuations in accuracy than FedAvg. FedSiKD records high accuracy even amidst high spikes due to the pronounced heterogeneity. Interestingly, at $\alpha = 0.1$, FedSiKD's accuracy shows a significant jump in the initial few rounds. This demonstrates that prior knowledge and clustering are effective for managing high non-i.i.d. data among clients, akin to the observed results on the MNIST dataset.  We conclude this discussion by stating that FedSiKD is highly suitable for environments with a high degree of Non-i.i.d data.

\begin{table*}
\centering
\caption{Test Loss on MNIST and HAR datasets for various Non-i.i.d levels.}
\begin{tabular}{lcccccccc}
\hline
& \multicolumn{4}{c}{\textbf{MNIST}} & \multicolumn{4}{c}{\textbf{HAR}} \\
\cline{2-9}
\textbf{Algorithm} & \multicolumn{2}{c}{$\alpha = 0.1$} & \multicolumn{2}{c}{$\alpha = 0.5$} & \multicolumn{2}{c}{$\alpha = 0.1$} & \multicolumn{2}{c}{$\alpha = 0.5$} \\
                   & 1st Round & Last Round & 1st Round & Last Round & 1st Round & Last Round & 1st Round & Last Round \\ \hline
\textbf{FedSiKD}            & \textbf{1.45}      &    \textbf{2.39}      & \textbf{0.59}      &  2.39       & 3.14      & 2.79       & \textbf{0.96}      & 5.39       \\
RandomCluster      & 1.93      & 4.94       & 0.63      & 2.19       & 2.00      & 9.22       & 1.00      & 13.28      \\
FL+HC              & 6.48       & 6.43       & 4.76      & 4.74       & 2.74      & 2.54       & 1.78      & 1.78       \\
FedAvg             & 7.67      & 33.25      & 1.17  & 46.61 & 4.13      & 13.99      & 2.73      & 88.66      \\ \hline
& \multicolumn{2}{c}{$\alpha = 1.0$} & \multicolumn{2}{c}{$\alpha = 2.0$} & \multicolumn{2}{c}{$\alpha = 1.0$} & \multicolumn{2}{c}{$\alpha = 2.0$} \\
                   & 1st Round & Last Round & 1st Round & Last Round & 1st Round & Last Round & 1st Round & Last Round \\ \hline
\textbf{FedSiKD}            & 0.48      & 1.41       & \textbf{0.37}      & 1.79       & 1.19      & 5.42       & 0.87      & 5.79       \\
RandomCluster      & 0.46      & 0.85       & 0.38      & 1.52       & 0.94      & 6.64       & 0.75      & 7.30       \\
FL+HC              & 4.29      & 4.31       & 5.89      & 5.89       & 2.22      & 2.50       & 1.81      & 2.31       \\
FedAvg             & 0.95  & 9.47  & 0.52 & 13.15   & 0.94      & 22.72      & 0.95      & 25.64      \\ \hline
\end{tabular}

\label{tab:fl_methods_comparison}
\end{table*}

In Table \ref{tab:fl_methods_comparison}, the test loss on \textbf{MINST} datasets for various Non-i.i.d levels is presented.
In extremely skewed data where $\alpha$ is 0.1, FedSiKD shows a significant difference from all other bases, almost 3\% better than the closest algorithm, RandomCluster. In all other cases, FedSiKD performs better than FL+HC and FedAvg, but slightly less than RandomCluster. Next, we discuss the test loss on \textbf{HAR} datasets for various Non-i.i.d levels. For all cases, FedSiKD performs better than RandomCluster and FedAvg. FL+HC shows better results than FedSiKD in all cases with a 2 to 3\% lead, except at low $\alpha$ levels, which show very similar results. Thus, we will analyze more results to observe the performance of all algorithms in the first five rounds in the next discussion.

\subsection{Few-Rounds Federated Learning}
In this section, we examine the effectiveness of FedSiKD in achieving superior performance with only a few rounds of communication in terms of accuracy and loss. Tables \ref{tab:mnist_shot} and \ref{tab:mnist_mnist_shot1} present the results from the first five rounds on the \textbf{MNIST} dataset against various values of $\alpha$. These results clearly demonstrate that FedSiKD outperforms other baseline algorithms. For instance, with $\alpha = 0.1$, there is a notable increase in test accuracy using FedSiKD, rising from 51\% to almost 75\%. This indicates that FedSiKD handles Non-i.i.d data very effectively, while other methods either improve slowly or show no significant change in performance. Similarly, in terms of test loss, when $\alpha = 0.1$, FedSiKD starts at 1.45 and decreases to 1 by the fifth round. In contrast, the test loss for other algorithms either increases or shows minimal improvement. The statistical data received from each client aids the global model in clustering clients based on similarity, which in turn helps to accelerate convergence, leading to improved accuracy and reduced loss. Similarly, at $\alpha$ 0.5 and 1.0, FedSiKD demonstrates robust and fast convergence, adapting to the data heterogeneity with significant improvements toward accuracy and reduction in loss in just a few rounds. At $\alpha$ = 2.0, FedSiKD shows better results than FL+HC and FedAvg, and is slightly less effective than RandomCluster, which shows the best performance with a difference of less than 1\% compared to FedSiKD.

\begin{table*}[h]
\centering
\caption{Comparison of accuracy and loss on the MNIST dataset for $\alpha$ = 0.1 and $\alpha$ = 0.5 over the first five rounds.}
\begin{tabular}{ccccc|ccccc}
\hline
\multicolumn{5}{c|}{\textbf{MNIST $\alpha = 0.1$}} & \multicolumn{5}{c}{\textbf{MNIST $\alpha = 0.5$}} \\
\hline
\textbf{Round} & \textbf{FL+HC} & \textbf{FedSiKD} & \textbf{RandomCluster} & \textbf{FedAvg}&\textbf{Round} & \textbf{FL+HC} & \textbf{FedSiKD} & \textbf{RandomCluster} & \textbf{FedAvg} \\
\hline
\multicolumn{5}{c|}{\textbf{Accuracy}} & \multicolumn{5}{c}{\textbf{Accuracy}} \\
1 & 57.53\% & 51.57\% & 42.80\% &37.66\% & 1 & 80.36\% & 83.01\% & 81.40\% & 71.40\%\\
2 & 56.71\% & 66.44\% & 61.83\% & 39.60\%&2 & 79.97\% & 83.68\% & 83.90\% &69.79\% \\
3 & 57.53\% & 71.42\% & 63.03\% &37.41\% &3 & 79.92\% & 86.32\% & 85.18\% &68.77\%\\
4 & 58.11\% & 74.18\% & 64.81\% &37.97\% &4 & 79.81\% & 86.47\% & 86.03\% &70.81\%\\
5 & 57.57\% & \textbf{74.83}\% & 65.18\% & 37.79\%& 5 & 80.13\% & \textbf{87.36}\% & 85.70\% &67.77\%\\
\hline
\multicolumn{5}{c|}{\textbf{Loss}} & \multicolumn{5}{c}{\textbf{Loss}} \\
1 & 6.48 & 1.45 & 1.93 &7.67 &1 & 4.76 & 0.59 & 0.63& 1.17\\
2 & 6.59 & 1.27 & 2.00 &10.10 &2 & 4.90 & 0.55 & 0.60& 1.49\\
3 & 6.82 & 1.07 & 2.53 &12.29 &3 & 4.96 & 0.49 & 0.59& 2.33\\
4 & 6.48 & 1.04 & 2.66 & 14.65&4 & 4.93 & 0.48 & 0.60 &2.52\\
5 & 6.44 & \textbf{1.02} & 3.05 &18.65 &5 & 4.82 & \textbf{0.47} & 0.65& 3.23\\
\hline
\end{tabular}

\label{tab:mnist_shot}
\end{table*}

\begin{table*}[ht]
\centering
\caption{Comparison of accuracy and loss on the MNIST dataset for $\alpha$ = 1.0 and $\alpha$ = 2.0 over the first five rounds}
\begin{tabular}{ccccc|ccccc}
\hline
\multicolumn{5}{c|}{\textbf{MNIST $\alpha = 1.0$}} & \multicolumn{5}{c}{\textbf{MNIST $\alpha = 2.0$}} \\
\hline
\textbf{Round} & \textbf{FL+HC} & \textbf{FedSiKD} & \textbf{RandomCluster} & \textbf{FedAvg} & \textbf{Round} & \textbf{FL+HC} & \textbf{FedSiKD} & \textbf{RandomCluster} & \textbf{FedAvg} \\
\hline
\multicolumn{5}{c|}{\textbf{Accuracy}} & \multicolumn{5}{c}{\textbf{Accuracy}} \\
1 & 85.87\% & 85.65\% & 86.18\% & 76.68\% & 1 & 86.42\% & 89.30\% & 89.41\% & 86.09\% \\
2 & 85.24\% & 89.02\% & 88.92\% & 75.01\% & 2 & 86.40\% & 89.94\% & 90.89\% & 86.27\% \\
3 & 85.56\% & 88.29\% & 88.98\% & 74.72\% & 3 & 86.52\% & 89.96\% & 90.88\% & 84.51\% \\
4 & 84.95\% & 87.86\% & 88.50\% & 74.24\% & 4 & 86.53\% & 90.01\% & 90.47\% & 83.71\% \\
5 & 85.10\% & \textbf{88.70}\% & 88.12\% & 74.81\% & 5 & 86.02\% & 90.03\% & 90.67\% & 85.60\% \\
\hline
\multicolumn{5}{c|}{\textbf{Loss}} & \multicolumn{5}{c}{\textbf{Loss}} \\
1 & 4.29 & 0.48 & 0.46 & 0.95 & 1 & 5.89 & 0.37 & 0.38 & 0.52 \\
2 & 4.43 & 0.38 & 0.39 & 1.47 & 2 & 5.63 & 0.37 & 0.35 & 0.70 \\
3 & 4.25 & 0.43 & 0.41 & 1.98 & 3 & 5.69 & 0.38 & 0.36 & 0.98 \\
4 & 4.37 & 0.45 & 0.43 & 2.40 & 4 & 5.64 & 0.39 & 0.38 & 1.12 \\
5 & 4.34 & \textbf{0.43} & 0.45 & 2.69 & 5 & 5.79 & 0.41 & 0.39 & 1.31 \\
\hline
\end{tabular}

\label{tab:mnist_mnist_shot1}
\end{table*}

A comparison of accuracy and loss on the \textbf{HAR} dataset for various $\alpha$ values throughout the first five rounds is shown in Tables \ref{tab:har_shot} and \ref{tab:har_shot1}. There is a notable increase in accuracy in the accuracy test for $\alpha = 0.1$, increasing from 31\% in the first round to 55\% in the fifth round. This significant increase suggests a great capacity for adaptation and learning from non-i.i.d. data. Using FedSiKD results in a decrease in test loss from 3.14\% to 2.06\% at $\alpha = 0.1$. Comparably, the test loss for $\alpha = 0.5$ drops from 0.96\% to 0.77\%, whereas the test loss for other methods either rises or demonstrates no improvement.
 
At $\alpha = 1.0$, FedSiKD surpasses FL+HC and FedAvg in terms of accuracy increases in the first five rounds. The accuracy of RandomCluster starts strong, but its growth throughout the rounds is less than that of FedSiKD. FedSiKD outperforms FL+HC and FedAvg at $\alpha = 2.0$, maintaining the lead from the start and through rounds one to five. RandomCluster performs similarly to FedSiKD but starts with higher initial accuracy. Finally, at  $\alpha$ = 1.0, and 2.0, the reduction in loss observed with FedSiKD indicates a positive trend, demonstrating an improvement in the model's predictive capabilities, while the loss associated with all other algorithms is increasing. We conclude that FedSiKD rapidly converges toward high accuracy, especially in cases of significant Non-i.i.d. data variance.

\subsection{Discussion Summary}
\textbf{Cluster-based Approach:} TThe findings show that when a global model is clustered with clients that have similar statistical distributions, it performs better overall. As a result, models become more precise and broadly applicable.

\textbf{Rapid Convergence:} The result show that FedSiKD achieves better results with fewer communication rounds. This suggests that the initial model distributed to clients efficiently captures knowledge between them
\textbf{Handling Non-i.i.d:} non-independent and identically distributed  (Non-i.i.d data) distribution is a significant challenge in Federated Learning (FL), and FedSiKD outperforms other base algorithms, especially in highly skewed data distributions. This indicates that prior knowledge of data distribution statistics effectively addresses this challenge.
\textbf{Resource-constrained:} FedSiKD leverages Knowledge
Distillation (KD) between clients, with the teacher being the large base model and students being built with fewer layers. These students effectively learn from knowledge transfers from the teacher to students
\textbf{Efficient Communication and Security:} The use of cluster mechanisms allows FedSiKD to run fewer communication rounds, reducing overhead and bandwidth usage. Where a higher number of communications may pose a risk of model exploitation.

\begin{table*}[h]
\centering
\caption{Comparison of accuracy and loss on the HAR dataset for $\alpha$ = 0.1 and $\alpha$ = 0.5 over the first five rounds.}
\begin{tabular}{ccccc|ccccc}
\hline
\multicolumn{5}{c|}{\textbf{HAR $\alpha = 0.1$}} & \multicolumn{5}{c}{\textbf{HAR $\alpha = 0.5$}} \\
\hline
\textbf{Round} & \textbf{FL+HC} & \textbf{FedSiKD} & \textbf{RandomCluster} & \textbf{FedAvg} & \textbf{Round} & \textbf{FL+HC} & \textbf{FedSiKD} & \textbf{RandomCluster} & \textbf{FedAvg} \\
\hline
\multicolumn{5}{c|}{\textbf{Accuracy}} & \multicolumn{5}{c}{\textbf{Accuracy}} \\
1 & 55.72\% & 31.01\% & 49.27\% & 50.93\% & 1 & 53.92\% & 66.41\% & 60.81\% & 58.91\% \\
2 & 50.56\% & 44.04\% & 48.90\% & 51.58\% & 2 & 59.04\% & 72.51\% & 61.32\% & 61.98\% \\
3 & 50.46\% & 46.45\% & 61.05\% & 50.29\% & 3 & 55.21\% & 77.37\% & 65.39\% & 61.18\% \\
4 & 49.85\% & 56.57\% & 55.85\% & 54.84\% & 4 & 58.26\% & 81.07\% & 65.52\% & 60.98\% \\
5 & 57.89\% & 55.55\% & 60.10\% & 52.49\% & 5 & 58.50\% & \textbf{81.61}\% & 71.67\% & 60.91\% \\
\hline
\multicolumn{5}{c|}{\textbf{Loss}} & \multicolumn{5}{c}{\textbf{Loss}} \\
1 & 2.74 & 3.14 & 2.00 & 4.13 & 1 & 1.78 & 0.96 & 1.00 & 2.73 \\
2 & 2.67 & 1.73 & 3.06 & 4.06 & 2 & 1.70 & 0.87 & 1.36 & 3.02 \\
3 & 2.51 & 2.48 & 2.08 & 4.26 & 3 & 1.78 & 0.76 & 1.20 & 4.86 \\
4 & 2.68 & 1.49 & 2.98 & 5.32 & 4 & 1.68 & 0.72 & 1.53 & 6.57 \\
5 & 2.61 & \textbf{2.06} & 3.34 & 6.90 & 5 & 1.64 & \textbf{0.77} & 1.17 & 9.05 \\
\hline
\end{tabular}

\label{tab:har_shot}
\end{table*}

\begin{table*}[ht]
\centering
\caption{Comparison of accuracy and loss on the HAR dataset for $\alpha$ = 1.0 and $\alpha$ = 2.0 over the first five rounds.}
\begin{tabular}{ccccc|ccccc}
\hline
\multicolumn{5}{c|}{\textbf{HAR $\alpha = 1.0$}} & \multicolumn{5}{c}{\textbf{HAR $\alpha = 2.0$}} \\
\hline
\textbf{Round} & \textbf{FL+HC} & \textbf{FedSiKD} & \textbf{RandomCluster} & \textbf{FedAvg} & \textbf{Round} & \textbf{FL+HC} & \textbf{FedSiKD} & \textbf{RandomCluster} & \textbf{FedAvg} \\
\hline
\multicolumn{5}{c|}{\textbf{Accuracy}} & \multicolumn{5}{c}{\textbf{Accuracy}} \\
1 & 59.82\% & 59.96\% & 63.35\% & 63.01\% & 1 & 55.21\% & 67.80\% & 71.97\% & 63.66\% \\
2 & 59.01\% & 66.81\% & 71.94\% & 65.52\% & 2 & 47.17\% & 73.60\% & 74.48\% & 63.73\% \\
3 & 59.79\% & 70.92\% & 72.72\% & 63.76\% & 3 & 50.29\% & 76.18\% & 79.67\% & 66.71\% \\
4 & 60.26\% & 74.75\% & 77.67\% & 64.51\% & 4 & 50.70\% & 78.22\% & 81.54\% & 68.95\% \\
5 & 59.11\% & 75.57\% & 77.47\% & 56.77\% & 5 & 51.00\% & 79.13\% & 82.83\% & 66.37\% \\
\hline
\multicolumn{5}{c|}{\textbf{Loss}} & \multicolumn{5}{c}{\textbf{Loss}} \\
1 & 2.22 & 1.19 & 0.94 & 0.94 & 1 & 1.81 & 0.87 & 0.75 & 0.95 \\
2 & 2.56 & 1.16 & 0.91 & 1.04 & 2 & 2.80 & 0.74 & 0.77 & 1.46 \\
3 & 2.53 & 1.17 & 1.06 & 1.25 & 3 & 2.51 & 0.76 & 0.74 & 1.21 \\
4 & 2.44 & 1.11 & 0.99 & 1.61 & 4 & 2.51 & 0.71 & 0.77 & 1.58 \\
5 & 2.79 & 1.15 & 0.97 & 1.89 & 5 & 2.29 & 0.79 & 0.76 & 1.76 \\
\hline
\end{tabular}

\label{tab:har_shot1}
\end{table*}

\section{Limitations and Future work}
In this research, we assume that each client shares their dataset distribution using a private filter, such as Differential Privacy. Applying a trade-off between noise and accuracy is not within the scope of this paper. FedSiKD performs clustering at the initial stage, once data distribution information is collected from clients. Dynamic clustering could be an interesting direction, as it involves balancing different factors such as privacy, overhead, and accuracy. Moreover, in the future work we fine-tuning teacher-student models to enhance the proposed work performance. 
\section{Conclusion}

This paper presents FedSiKD, an approach that addresses the non-i.i.d. nature of client data and resource constraints by integrating knowledge distillation (KD) into a similarity-based federated learning system. FedSikd handles resource constraints, improves optimization efficiency in highly skewed datasets, and ensures fast learning convergence. FedSiKD outperforms state-of-the-art algorithms in terms of accuracy and shows impressive early-stage learning capabilities, that cloud be suitable to highly sensitive party who hesitate to adopt FL.

\bibliographystyle{IEEEtran}
\bibliography{main}

\end{document}